\documentclass[letterpaper, 10 pt, journal, twoside]{IEEEtran}
\usepackage{amsmath,amsfonts}
\usepackage{algorithm}
\usepackage{amssymb}
\usepackage{array}
\usepackage[caption=false,font=footnotesize,labelfont=rm,textfont=rm]{subfig}
\usepackage{textcomp}
\usepackage{color,xcolor}

\usepackage{stfloats}
\usepackage{url}
\usepackage{verbatim}
\usepackage{graphicx}
\usepackage{cite}
\usepackage{multirow}

\usepackage{algorithmicx}   
\usepackage{algpseudocode}  
\usepackage{tabularx}
\usepackage{booktabs}
\usepackage{enumerate}
\usepackage{breqn}
\usepackage{indentfirst}
\usepackage{bm}
\usepackage{mathtools}
\usepackage{adjustbox}
\usepackage{colortbl}
\usepackage{threeparttable}
\usepackage{arydshln}

\begin{document}
\title{R-VoxelMap: Accurate Voxel Mapping with Recursive Plane Fitting for Online LiDAR Odometry}

\author{Haobo Xi, Shiyong Zhang*,~\IEEEmembership{Member,~IEEE}, Qianli Dong,~\IEEEmembership{Student Member,~IEEE}, Yunze Tong,~\IEEEmembership{Student Member,~IEEE}, Songyang Wu, Jing Yuan,~\IEEEmembership{Member,~IEEE}, Xuebo Zhang,~\IEEEmembership{Senior Member,~IEEE}
\thanks{Manuscript received: August 23, 2025; Revised: November 3, 2025; Accepted: December 17, 2025.}
\thanks{
This paper was recommended for publication by Javier Civera upon evaluation of the Associate Editor and Reviewers' comments.
This work was supported in part by National Natural Science Foundation of China under Grant Number 62303249, in part by the China Postdoctoral Science Foundation under Grant Number 2024M751526, and in part by the Beijing-Tianjin-Hebei Fundamental Research Cooperation Project under Grant Number 24JCZXJC00390.
(Corresponding author: Shiyong Zhang.)}
\thanks{All authors are with the College of Artificial Intelligence, Institute of Robotics and Automatic Information System, and the Tianjin Key Laboratory of Intelligent Robotics, Nankai University, Tianjin 300350, China (e-mail: xihaobo@mail.nankai.edu.cn; zhangshiyong@nankai.edu.cn).}
\thanks{Digital Object Identifier (DOI): see top of this page.}
}

\markboth{IEEE Robotics and Automation Letters. Preprint Version. Accepted December, 2025}%
{Xi \MakeLowercase{\textit{et al.}}: R-VoxelMap: Accurate Voxel Mapping with Recursive Plane Fitting for Online LiDAR Odometry}

\maketitle

\begin{abstract}
This paper proposes R-VoxelMap, a novel voxel mapping method that constructs accurate voxel maps using a geometry-driven recursive plane fitting strategy to enhance the localization accuracy of online LiDAR odometry.
VoxelMap and its variants typically fit and check planes using all points in a voxel, which may lead to plane parameter deviation caused by outliers, over segmentation of large planes, and incorrect merging across different physical planes.
To address these issues, R-VoxelMap utilizes a geometry-driven recursive construction strategy based on an outlier detect-and-reuse pipeline.
Specifically, for each voxel, accurate planes are first fitted while separating outliers using random sample consensus (RANSAC). The remaining outliers are then propagated to deeper octree levels for recursive processing, ensuring a detailed representation of the environment.
In addition, a point distribution-based validity check algorithm is devised to prevent erroneous plane merging.
Extensive experiments on diverse open-source LiDAR(-inertial) simultaneous localization and mapping (SLAM) datasets validate that our method achieves higher accuracy than other state-of-the-art approaches, with comparable efficiency and memory usage.
Code will be available on GitHub.\footnote{\url{https://github.com/NKU-MobFly-Robotics/R-VoxelMap}}
\end{abstract}


\section{Introduction}
The rapid development of 3D light detection and ranging (LiDAR) technology has greatly accelerated the adoption of LiDAR(-inertial) odometry across diverse applications.
Due to its high accuracy, this technology has been widely utilized and extensively researched over the past decade. 
To support accurate and real-time state estimation, mainstream LiDAR odometry approaches increasingly rely on map structures that can accurately represent the environment while supporting efficient maintenance and registration. 
As the most classic map structure, k-d tree enables point cloud registration via k-nearest neighbor (KNN) search and is widely used in LiDAR(-inertial) odometry\cite{loam, lio-sam,lins,fast-lio}. 
Fast-LIO2 \cite{fast-lio2} proposes an incremental k-d tree, enabling more efficient map updates.
However, such tree-based methods still face time-consuming KNN search and complex tree updates.
To further improve efficiency, voxel-based methods have been proposed and become the mainstream\cite{faster-lio, voxelmap, yaohekai}. 
These methods partition space into voxels and apply spatial hashing to enable fast point matching and registration, significantly enhancing efficiency while reducing the implementation complexity of map maintenance.

\begin{figure}[t]
    \centering
    \includegraphics[width=0.48\textwidth]{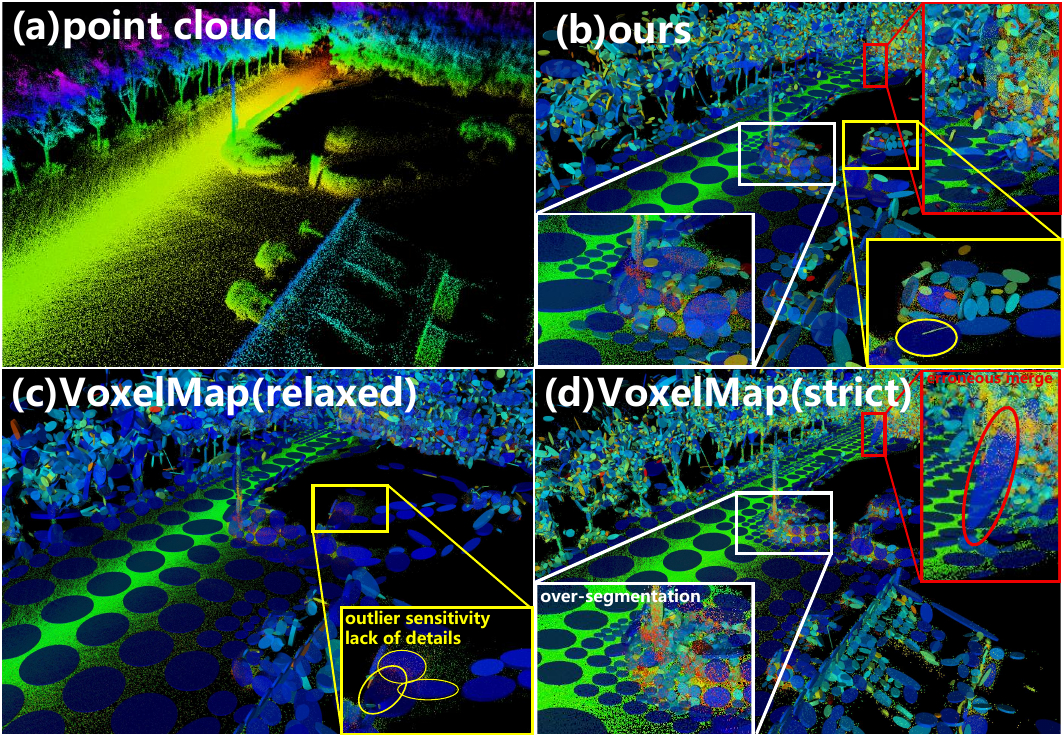}
    \caption{
       VoxelMap plane features suffers from outlier sensitivity (c, yellow-marked planes are erroneously shifted due to surrounding outliers), over-segmentation (d, white-boxed planar features are overly segmented, resulting in large uncertainty), and incorrect plane merging
       (d, red-marked regions show erroneous plane merging, generating large planar structures in areas that are not physically planar), all of which reduce localization accuracy.
       Our method effectively addresses these limitations.
    }  
    \label{fig:first}
    \vspace{-0.5cm}  
\end{figure}

As one of the most widely adopted voxel-based map structures, VoxelMap \cite{voxelmap} enables odometry methods to achieve improved accuracy and efficiency using probabilistic plane features.
However, since it performs plane fitting using all points within a voxel, it inevitably suffers from two major limitations.
First, the fitted planes are sensitive to outliers, even a small number of outliers can significantly distort the estimated plane parameters, thereby degrading localization and mapping accuracy (as shown in Fig.~\ref{fig:first}(c)).
While stricter plane calculation thresholds help reduce outlier inclusion, they often lead to over-segmentation (as shown in Fig.~\ref{fig:first}(d)), resulting in fragmented planes with fewer points and slower uncertainty convergence, which is not conducive to accurate pose estimation.
To alleviate this issue, VoxelMap's variants\cite{voxelmap++,c3p-voxelmap} further merge coplanar planes to reconstruct larger planes with more point support and lower parameter uncertainty.
However, since plane fitting before merging still relies on all points within a voxel, the initially extracted small planes may exhibit parameter deviations, which can eventually result in incomplete merging or incorrect merging.
Second, points from different physical surfaces may be erroneously merged into the same plane, causing map distortion (as shown in Fig.~\ref{fig:first}(d)).

To address these challenges, we propose R-VoxelMap: a geometry-driven recursive plane fitting strategy for accurate voxel mapping.
Different from other variants, we do not follow a split-and-merge strategy.
Instead, we directly use the proposed outlier detect-and-reuse pipeline during the coarse-to-fine plane feature extraction, which effectively suppresses the influence of outliers.
By simultaneously ensuring accurate plane feature extracting and preserving multi-scale geometric information, the proposed method effectively enhances odometry accuracy.
Our contributions can be concluded as follows:
\begin{enumerate}
\item A novel plane feature voxel map structure and construction strategy, R-VoxelMap, which is constructed via geometry-driven recursive plane fitting.
By performing an outlier detect-and-reuse pipeline in each recursive iteration, R-VoxelMap effectively suppresses the influence of outliers, enables more accurate map representation and reduces plane over-segmentation.
\item A point distribution-based plane validity check algorithm that projects and clusters point cloud on the RANSAC-fitted plane to effectively prevent the incorrect merging of different physical planes. 
\item Extensive comparative experiments on multiple open-source LiDAR(-inertial) SLAM datasets demonstrate that our method achieves higher state estimation accuracy than other state-of-the-art approaches.
The code will be open-sourced to benefit the community.
\end{enumerate}

\begin{figure*}[ht]
   \vspace*{0.5\baselineskip} 
   \centering
   \includegraphics[width=\textwidth]{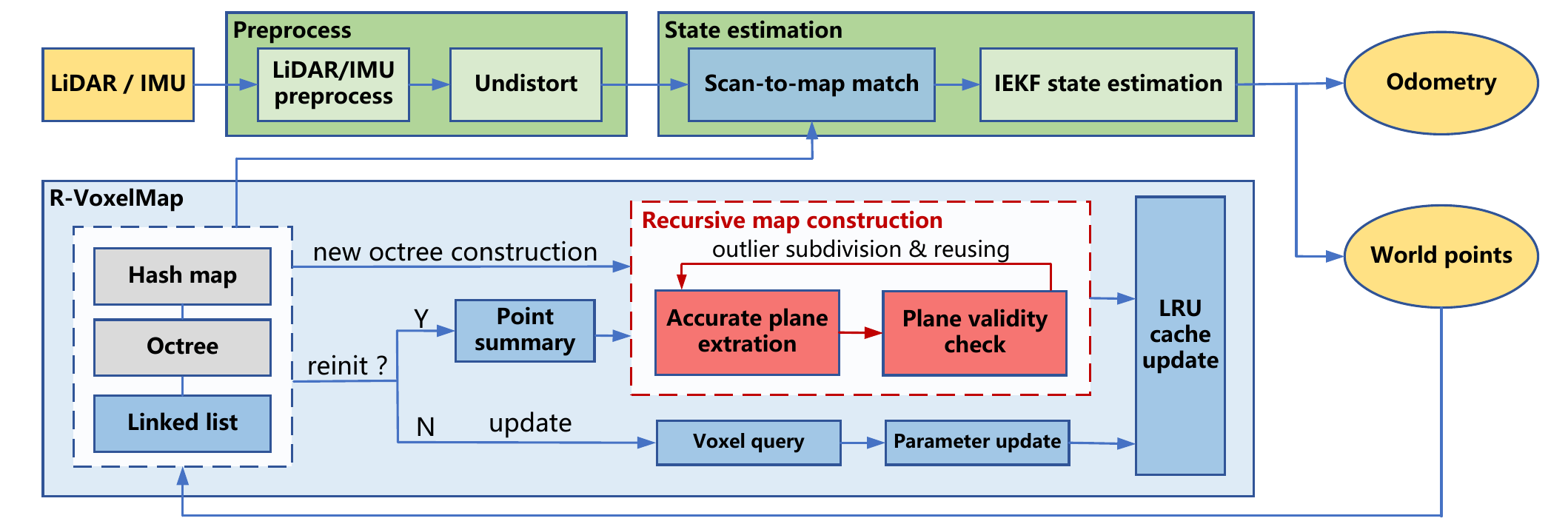}
   \caption{System overview of the LiDAR-based odometry using R-VoxelMap. 
   The red-highlighted modules represent the core of R-VoxelMap, a geometry-driven recursive map construction algorithm based on an outlier detect-and-reuse pipeline.
   The blue modules are specifically designed to support the fundamental functions of R-VoxelMap.
   }
   \label{fig:system}
   \vspace{-0.5cm}  
\end{figure*}

\section{Related Work}
The map structure for LiDAR odometry is required to support accurate and efficient point cloud registration, and can be categorized into tree-based and voxel-based approaches.
Based on these map structures, LiDAR odometry can perform accurate point cloud registration using iterative closest points (ICP)\cite{icp} or generalized ICP (GICP)\cite{gicp} algorithms.
Classic LiDAR odometry methods\cite{loam, lego-loam} and a range of LiDAR-inertial odometry systems\cite{lio-sam,lins,fast-lio}, 
organize plane and edge feature points using k-d tree and perform registration via ICP.
The efficiency of these methods is often constrained by the slowness of k-d tree searches and frequent tree reconstructions.
Fast-LIO2\cite{fast-lio2} uses a novel incremental k-d tree to reduce the overhead of frequent reconstructions. But its performance is still constrained by KNN search and rebalancing overhead.
Faster-LIO\cite{faster-lio} replaces the tree-based map structures with incremental voxels, which improves computational efficiency while maintaining localization accuracy. 
However, these point cloud-based mapping algorithms typically treat matched points as accurate representations of the environment, without accounting for the uncertainty introduced by sensor measurements and pose estimation.

VoxelMap\cite{voxelmap} employs an adaptive-size, coarse-to-fine voxel-based map structure based on hash map and octrees, where each voxel contains a probabilistic plane feature.
This structure supports accurate modeling of the environment while considering uncertainty.
Current VoxelMap variants typically improve odometry accuracy by merging coplanar plane features.
For example, VoxelMap++\cite{voxelmap++} reduces plane uncertainty and sensitivity to noise by merging potentially coplanar planes using a union-find algorithm.
C3P-VoxelMap\cite{c3p-voxelmap}, on the other hand, proposes an on-demand voxel merging strategy based on locality-sensitive hashing.
These works confirm that merging coplanar points to form more robust plane features enhances pose estimation accuracy.

However, all these VoxelMap-based methods \cite{voxelmap, voxelmap++, c3p-voxelmap} fit plane features using all points within each voxel, making them sensitive to outliers, which can distort the map.
Simply tightening the eigenvalue threshold may alleviate this issue but often leads to plane over-segmentation, increasing uncertainty and ultimately degrading odometry accuracy.
To this end, we propose R-VoxelMap, which enhances environment representation accuracy by recursively constructing map via an outlier detect-and-reuse pipeline.
Unlike previous methods that optimize the map representation through plane merging and are still susceptible to outlier interference,
our approach directly suppresses the influence of outliers during the plane extraction stage, yielding more accurate plane parameters and enabling the extraction of larger planes with greater point support,
thereby improving the overall odometry accuracy.

\begin{figure*}[ht]
   \centering
   \includegraphics[width=\textwidth]{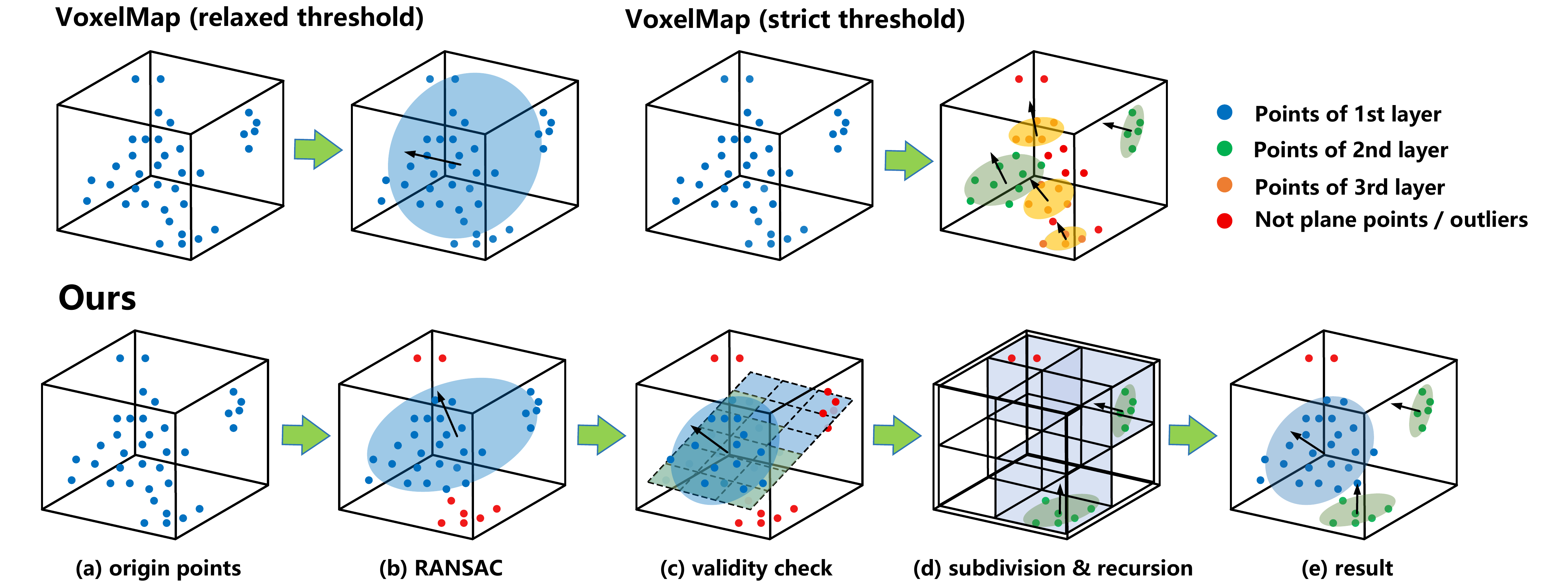}
   \caption{Comparison between R-VoxelMap construction (lower part) and VoxelMap construction (upper part).
   VoxelMap's plane parameters are sensitive to outliers under relaxed thresholds (upper left), while strict thresholds cause over-segmentation (upper right).
   Our recursive method uses outlier detect-and-reuse pipeline and plane validity check to effectively reduce these issues (see Algorithm~1).}
   \label{fig:ransac}
   \vspace{-0.5cm}  
\end{figure*}

\section{Methodology}
\subsection{System Overview}
R-VoxelMap is a collection of voxels, each associated with a probabilistic plane.
It is embedded into an iterated extended kalman filter (IEKF) framework for pose estimation, and the whole system architecture is shown in Fig.~\ref{fig:system}.
If there is no inertial measurement unit (IMU) available, the proposed method can be used as a standalone LiDAR odometry system with a constant velocity model.

The underlying structure of R-VoxelMap is a position-keyed hash table, where each key corresponds to an octree representing a low-resolution voxel.
Each voxel is further adaptively subdivided into smaller sub-voxels at varying resolutions based on point cloud distribution.
Unlike previous VoxelMap approaches, R-VoxelMap allows every node in the octree (not just leaf nodes) to store points and probabilistic planes.
Each voxel stores its points using a linked list to enable efficient partition and reuse.
When a new point cloud is received, R-VoxelMap constructs new octrees for newly observed voxels by applying RANSAC-based accurate plane extraction to reduce the impact of outliers and improve environmental representation accuracy.
The outliers are subdivided and processed recursively in the octree (see Sec.~\ref{sec:ransac}).
To prevent merging distinct physical planes, a point distribution-based plane validity check is performed, retaining only geometrically consistent planes (see Sec.~\ref{sec:check}).
For existing octrees, R-VoxelMap supports efficient octree update and reconstruction based on linked-list points organization (see Sec.~\ref{sec:update}).

\subsection{R-VoxelMap Construction}\label{sec:ransac}
To address VoxelMap's issue of outlier sensitivity and plane over-segmentation caused by using all points in a voxel for plane fitting via eigenvalue decomposition,
R-VoxelMap adopts a recursive map construction method based on an outlier detect-and-reuse pipeline.
This section provides a detailed description of this recursive construction.

Similar to VoxelMap, R-VoxelMap constructs probabilistic plane features using point cloud in world frame that simultaneously accounts for measurement uncertainty and state estimation uncertainty.
Taking the LiDAR odometry as an example, given a LiDAR point ${}^L\mathbf{p}_i$ with associated covariance $\Sigma_{{}^L\mathbf{p}_i}$ in the LiDAR frame,
and a frame pose estimated by odometry as ${}^W_L\mathbf{T} = ({}^W_L\mathbf{R}, {}^W_L\mathbf{t}) \in \mathit{SE}(3)$ with uncertainties $\Sigma_{{}^W_L\mathbf{R}}$ and $\Sigma_{{}^W_L\mathbf{t}}$,
the transformed point in the world frame and its uncertainty can be expressed as
\begin{equation}
   \begin{gathered}
{}^W\mathbf{p}_i = {}^W_L\mathbf{R} {}^L\mathbf{p}_i + {}^W_L\mathbf{t},\\
\Sigma_{{}^W\!\mathbf{p}_{i}}\! =\! {}_{L}^{W}\mathbf{R}\Sigma_{^L{\mathbf{p}_{i}}}{}^W_L\mathbf{R}^{T}\!+\!
                              {}_{L}^{W}\mathbf{R}\lfloor{}^{L}\mathbf{p}_{i}\rfloor_{\wedge}\Sigma_{\mathbf{R}}\lfloor^{L}\mathbf{p}_{i}\rfloor_{\wedge}^{T} {}_{L}^{W} \mathbf{R}^{T}\!+\!
                              \Sigma_{\mathbf{t}}.
\end{gathered}
\end{equation}
To construct the map structure, R-VoxelMap first calculates the low-resolution voxels that world points belong to, via hash map.
For each low-resolution voxel, if the number of points is enough for initialization, an octree is constructed through the proposed recursive construction process (see Algorithm~1).

\algrenewcommand\algorithmicrequire{\textbf{Input:}}
\algrenewcommand\algorithmicensure{\textbf{Output:}}

\begin{algorithm}
\caption{Recursive Octree Construction}
\begin{algorithmic}[1]
\State \textbf{Input:} Point cloud $\mathcal{P}_{\mathrm{in}}$, current octree node depth $D_{\mathrm{cur}}$
\State \textbf{Parameters:} Distance threshold $d_{th}$, inlier ratio threshold $p_{th}$, RANSAC iteration times $K$, minimum point threshold $N_{\mathrm{min}}$, octree max depth $D_{\mathrm{max}}$, plane threshold $\lambda_{\mathrm{th}}$
\State Initialize: $\mathrm{BestPlane}\ \pi^{*} \gets \emptyset$, $\mathrm{MaxInliers} \gets 0$
\State Initialize: $\mathrm{Inlier\ set}\ \mathcal{I} \gets \emptyset$, $\mathrm{Outlier\ set}\  \mathcal{O} \gets \emptyset$
   
   \State $(\mathcal{I},\mathcal{O}) \gets$ RANSAC($\mathcal{P}_{\mathrm{in}}$, $d_{th}$, $K$)
   \If{$\frac{|\mathcal{I}|}{|\mathcal{P}_{\mathrm{in}}|} > p_{th}$}
       \State $(\pi^{*},\lambda_{\mathrm{min}}) \gets$  Fit plane from $\mathcal{I}$ 
       \If{$\lambda_{\min} < \lambda_{\mathrm{th}}$}
       \State $(\mathcal{I}_{\mathrm{valid}}, \mathcal{O}_{\mathrm{new}}) \gets$ PlaneValidityCheck($\pi^*, \mathcal{I}$)
       \State $\mathcal{O} \gets  \mathcal{O} \cup \mathcal{O}_{\mathrm{new}}$
       \If{$|\mathcal{I}_{\mathrm{valid}}| > 0$}
          \State $\pi^{*}\gets$ Fit final plane using $\mathcal{I}_{\mathrm{valid}}$
       \EndIf
       \Else
          \State $\mathcal{O} \gets \mathcal{P}_{\mathrm{in}}$
       \EndIf
   \Else
       \State $\mathcal{O} \gets \mathcal{P}_{\mathrm{in}}$
   \EndIf
   
\If{$|\mathcal{O}| > 0$ and $D_{\mathrm{cur}} \le  D_{\mathrm{max}}$}
\State Subdivide $\mathcal{O}$ into octree subvoxels $\{V_k\}$
\ForAll{$V_k$ such that $|\mathcal{P}_{\mathrm{in},V_k}| \geq N_{\mathrm{min}}$}
\State Recursively call algorithm on $V_k$ with $D_{\mathrm{cur}} + 1$
\EndFor 
\EndIf
\end{algorithmic}
\end{algorithm}

Specifically, for all point cloud within each low-resolution voxel, the RANSAC\cite{ransac} plane fitting algorithm
is first applied to separate outliers (denoted as $\mathcal{O}$) from inliers (denoted as $\mathcal{I}$).
If the proportion of inliers relative to the total number of points exceeds the predefined threshold, all inliers ${}^W\mathbf{p}_i \in \mathcal{I}$ are then used to estimate the plane parameters.
First, the center point $\mathbf{q}$ and covariance matrix $\mathbf{A}$ are computed as follows:
\begin{equation}
   \begin{gathered}
      \mathbf{q}=\frac{1}{|\mathcal{I}|}\sum_{{}^W\mathbf{p}_i \in \mathcal{I}}{}^W\mathbf{p}_{i} , \\
      \mathbf{A}=\frac{1}{|\mathcal{I}|}\sum_{{}^W\mathbf{p}_i \in \mathcal{I}}{}^W\mathbf{p}_{i}{}^W\mathbf{p}_{i}^{T}-\mathbf{q}\mathbf{q}^{T} = \frac{1}{|\mathcal{I}|}\mathbf{S}-\mathbf{q}\mathbf{q}^{T}.
   \end{gathered}
\end{equation}
$\mathbf{S}=\sum_{{}^W\mathbf{p}_i \in \mathcal{I}}{}^W\mathbf{p}_{i}{}^W\mathbf{p}_{i}^{T}$ is stored for incremental update.
Then, an eigenvalue decomposition of the matrix $\mathbf{A}$ is performed to obtain its eigenvalues and the corresponding eigenvector matrix $\mathbf{U}$.
The eigenvector associated with the smallest eigenvalue $\lambda_{\text{min}}$ is used as the normal vector $\mathbf{n}$ of the plane.
If $\lambda_{\text{min}}$ satisfies the plane threshold, the plane validity check (Sec.~\ref{sec:check}) is conducted to further remove points that do not conform to current plane.
Only one geometrically consistent plane and its inliers (i.e., $\mathcal{I}_{\text{valid}}$) are retained, while all other points are classified as outliers.
Final plane parameters are calculated from $\mathcal{I}_{\text{valid}}$ and stored in the current octree node.
In consideration of noise, the actual plane parameters can be expressed as
\begin{equation}
   \begin{aligned}
      [\mathbf{n}_{gt}^T,\mathbf{q}_{gt}^T]^T&=\mathbf{f}(\mathbf{p}_{1}+{\delta}_{\mathbf{p}_1},\mathbf{p}_{2}+{\delta}_{\mathbf{p}_2},...,\mathbf{p}_{N}+{\delta}_{\mathbf{p}_N})\\
      &\approx[{\mathbf{n}}^T,{\mathbf{q}}^T]^T+\sum_{\mathbf{p}_i \in \mathcal{I}_{\text{valid}}}\frac{\partial\mathbf{f}}{\partial\mathbf{p}_i}\delta_{\mathbf{p}_i}
   \end{aligned}
\end{equation}
where ${}^W\mathbf{p}_i$ is simplified as $\mathbf{p}_i$. The Jacobian of the parameter $(\mathbf{n},\mathbf{q})$ with respect to each world-frame point $\mathbf{p}_i$ is computed according to~\cite{voxelmap} as
\begin{equation}
   \begin{aligned}
      \frac{\partial\mathbf{n}}{\partial\mathbf{p}_{i}}=\mathbf{U}\left[\begin{array}{c}\mathbf{F}_{1}(\mathbf{p}_i)\\\mathbf{F}_{2}(\mathbf{p}_i)\\\mathbf{F}_{3}(\mathbf{p}_i)\end{array}\right],
      \frac{\partial\mathbf{q}}{\partial\mathbf{p}_{i}}=\mathrm{diag}(\frac{1}{N},\frac{1}{N},\frac{1}{N}),
   \end{aligned}
\end{equation}
\begin{equation}
   \begin{aligned}
      \mathbf{F}_{m}(\mathbf{p}_i)=\left\{\begin{array}{cc}\frac{(\mathbf{p}_{i}-\mathbf{q})^{T}}{N(\lambda_3-\lambda_m)}\left(\mathbf{u}_m\mathbf{n}^T+\mathbf{n}\mathbf{u}_m^T\right)&,m\neq3,\\\mathbf{0}_{1\times3}&,m=3.\end{array}\right.
   \end{aligned}
\end{equation}
where $\lambda_{3}$ is the minimum eigenvalue. Accordingly, the uncertainty of the plane parameters can be expressed by the covariance matrix as follows:
\begin{equation}
   \Sigma_{\mathbf{n},\mathbf{q}}=\sum_{\mathbf{p}_i \in \mathcal{I}_{\text{valid}}}\frac{\partial\mathbf{f}}{\partial\mathbf{p}_{i}}\Sigma_{\mathbf{p}_{i}}\frac{\partial\mathbf{f}}{\partial\mathbf{p}_{i}}^{T},
   \frac{\partial\mathbf{f}}{\partial\mathbf{p}_{i}}={\left[{\frac{\partial\mathbf{n}}{\partial\mathbf{p}_{i}}^T},{\frac{\partial\mathbf{q}}{\partial\mathbf{p}_{i}}^T}\right]}^T
\end{equation}
$\Sigma_{\mathbf{p}_{i}}$ is the covariance matrix of the world point $\mathbf{p}_i$, computed by (1).
The covariance matrix is also stored in current octree node for subsequent point-to-plane matching.

After the plane fitting process, all outlier points are subdivided in the octree, and the points in each sub-voxel are treated as new candidate points (i.e., $\mathcal{P}_{\text{in},V_k}$).
Once their count exceeds the predefined threshold, the construction process is recursively applied on sub-voxels.
This recursive process continues until all points are classified as inliers, or the maximum octree depth is reached.

\subsection{Plane Validity Check}\label{sec:check}
\begin{figure}[]
   \vspace*{0.5\baselineskip} 
   \centering
   \includegraphics[width=0.5\textwidth]{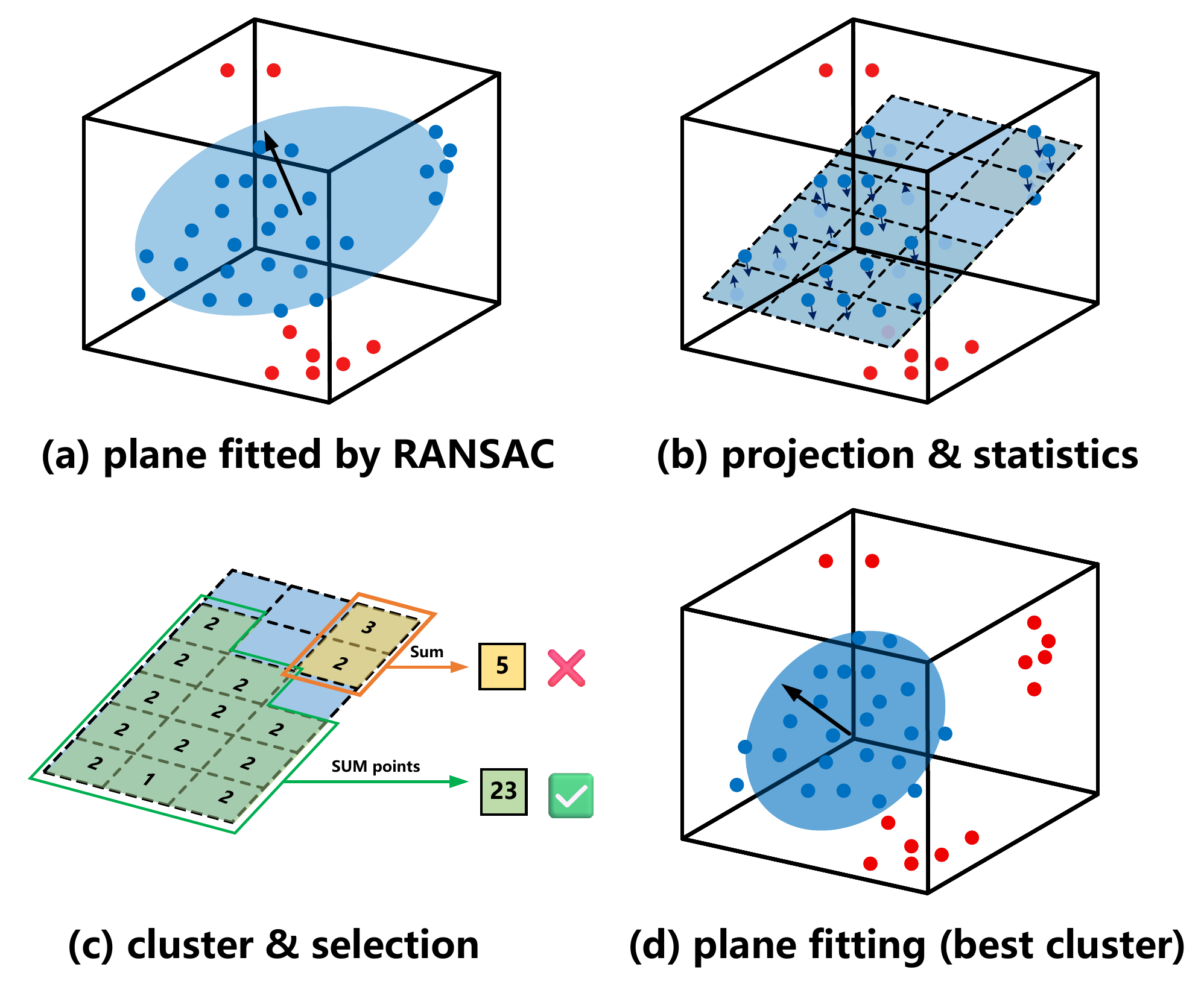}
   \caption{Illustration of the point distribution-based plane validity check.
   (a) Obtain the best candidate plane from RANSAC result.
   (b) Discretize the plane into 2D grids with a fixed resolution, project inlier points onto the plane and count the number of points in each grid.
   (c) Cluster occupied grids and compute the total number of points for each cluster.
   (d) Select the optimal cluster and refit the plane using its points.}
   \label{fig:check}
   \vspace{-0.5cm}  
\end{figure}

Although RANSAC can effectively excludes outliers, it may still erroneously merge points from different planes, especially when planes have similar but distinct normal vectors.
This leads to two main issues: (1) the estimated plane parameters are biased by points from other planes, resulting in inaccurate plane parameters and unreliable uncertainty;
(2) during matching, the odometry may incorrectly assume the presence of a plane in empty space (e.g., the space between two point clusters in Fig.~\ref{fig:check}(a)), causing false point-to-plane matching.
To address this issue, the plane validity check is performed to further remove outliers from the inlier set.

Specifically, the plane fitted by RANSAC is used as the reference plane for validity check.
The 2D space of the reference plane is uniformly discretized around the mean point $\mathbf{q}$, with a fixed resolution of $r=\text{voxel\_size}/n$,
where $n$ is a hyperparameter that controls the sensitivity of the validity check.
All inlier points are projected onto the reference plane (Fig.~\ref{fig:check}(b)), and their corresponding grids are computed via
\begin{equation}
   (x_i, y_i) = \left( \left\lfloor \frac{(\mathbf{p}_i - \mathbf{q}){}^{T} \mathbf{u}_1}{r} \right\rfloor, \; \left\lfloor \frac{(\mathbf{p}_i - \mathbf{q}){}^{T} \mathbf{u}_2}{r} \right\rfloor \right)
\end{equation}
where $\mathbf{u}_1$ and $\mathbf{u}_2$ are the two eigenvectors of the covariance matrix $\mathbf{A}$ corresponding to the two largest eigenvalues.
The number of points in each occupied grid is counted during projection.
All occupied grid are clustered using a depth-first search based region-growing algorithm, where grids connected via four-neighbor adjacency are grouped into the same cluster (Fig.~\ref{fig:check}(c)).
Each cluster accumulates the total number of its points, and the one containing the largest number of points is selected as the best cluster. (Fig.~\ref{fig:check}(d)).
The plane is considered valid if the point number of the best cluster satisfies
\begin{equation}
   \frac{|\mathcal{I}_{\text{best}}|}{|\mathcal{P}_{\text{in}}|} > p_{th}
\end{equation}
where $p_{th}$ is the minimum inlier ratio.
The final plane for this octree node is refitted using the valid points in the best cluster.
Otherwise, all planes are considered invalid and all points are classified as outliers.

\subsection{Map Update and Octree Reconstruction}\label{sec:update}
Since octree construction requires accessing all points in each voxel, rebuilding it for every new point is inefficient.
To address this, R-VoxelMap adopts an incremental update strategy.
When a new point is assigned to an existing octree, the system selects the nearest plane in it as the matching candidate and calculates the plane parameters incrementally as follows:
\begin{equation}
   \mathbf{q} = \frac{N}{N+1} \, \mathbf{q} + \frac{1}{N+1} \, \mathbf{p}_{\text{new}} ,\quad
   \mathbf{S} = \mathbf{S} + \mathbf{p}_{\text{new}} \, \mathbf{p}_{\text{new}}^{T}
\end{equation}
where $N$ is the number of plane points.
Using the updated $\mathbf{q}$ and $\mathbf{S}$, compute $\mathbf{A}$ via Eq.~(2) and obtain its smallest eigenvalue $\lambda_{\min}$.
If $\lambda_{\min}$ is smaller than the predefined threshold, the point is accepted and used to update the plane parameters.
Otherwise, the point is rejected and stored in the leaf node's non-plane point set for potential use in future.

However, only relying on incremental updates after the initial octree construction may result in inaccuracies, as the sparse point cloud at early stages may not sufficiently represent the environment, leading to suboptimal initialization of the octree and plane parameters.
Furthermore, as the sensor's viewpoint changes, previously mapped voxels may acquire new environmental information, which may not be captured in the current octree.
To address this, R-VoxelMap periodically reconstructs the octree after accumulating sufficient points.
Leveraging its linked-list-based point storage, R-VoxelMap accesses all points efficiently by splicing point linked lists within the octree, avoiding the time and memory overhead of data copying in vector-based methods.

A least recently used (LRU) cache, implemented with a hash table and a linked list, is employed to prevent excessive memory usage of the map in large-scale environments.
Low-resolution voxels are used as the basic elements managed by the cache.
After each map update, the LRU cache is refreshed by moving the updated voxels to the front of the queue while removing voxels that exceed the cache capacity.

\subsection{Point-to-Plane Matching for State Estimation}
In the scan-to-map matching process, all planes within the corresponding low-resolution voxel are treated as candidate planes, and the point-to-plane distance measurement equations are constructed following VoxelMap approach:
\begin{equation}
   \begin{aligned}
      d_{i}&=\mathbf{n}_i^T(^W\mathbf{p}_i-\mathbf{q}_i)\\
      &=(\mathbf{n}_{i}^{gt} \boxplus \boldsymbol{\delta}_{\mathbf{n}_{i}})^{T}[(^{W}\mathbf{p}_{i}^{gt}+\boldsymbol{\delta}_{{}^W{\mathbf{p}_{i}}})-\mathbf{q}_{i}^{gt}-\boldsymbol{\delta}_{\mathbf{q}_{i}}]\\
      &\approx{\mathbf{n}_i^{gt}}^T({}^W\mathbf{p}_i^{gt}-\mathbf{q}_i^{gt})+\mathbf{J}_{\mathbf{n}_i}\delta_{\mathbf{n}_i}+\mathbf{J}_{\mathbf{q}_i}\delta_{\mathbf{q}_i}+\mathbf{J}_{{}^W{\mathbf{p}_i}}\delta_{{}^W{\mathbf{p}_i}}\\
      &\sim\mathcal{N}(0,\sigma^2_{i}).
   \end{aligned}
\end{equation}
The variance of the measurement noise $\sigma^2_{i}$ can be computed as follow:
\begin{equation}
   \begin{aligned}
      \sigma^2_{i}&=\mathbf{J}_{\mathbf{w}_{i}}\begin{bmatrix}\boldsymbol{\Sigma}_{\mathbf{n}_i,\mathbf{q}_i}&0\\0&\boldsymbol{\Sigma}_{^W\mathbf{p}_i}\end{bmatrix}\mathbf{J}_{\mathbf{w}_{i}}^{T},\\
      \mathbf{J}_{\mathbf{w}_{i}}&=\begin{bmatrix}\mathbf{J}_{\mathbf{n}_{i}},\mathbf{J}_{\mathbf{q}_{i}},\mathbf{J}_{^W{\mathbf{p}_{i}}}\end{bmatrix}=\begin{bmatrix}(^{W}\mathbf{p}_{i}-\mathbf{q}_{i})^{T},-\mathbf{n}_{i}^{T},\mathbf{n}_{i}^{T}\end{bmatrix}.
   \end{aligned}
\end{equation}
The probability that this point belongs to the current plane can be further calculated as
\begin{equation}
   p_{i} = \frac{1}{\sigma_{i}\sqrt{2\pi}} \exp\left(-\frac{d_{i}^{2}}{2\sigma^2_{i}}\right).
\end{equation}
Among all candidate planes that meet the distance threshold requirement, the one with the highest probability is selected as the final matching plane.
All final matched result is fused with prior information for state estimation via maximum a posteriori (MAP) inference.

\section{Experiments}
In this section, to validate the effectiveness of the proposed method, we conduct experiments on multiple datasets and compare our method with current state-of-the-art algorithms.
Specifically, the datasets used in our experiments include KITTI\cite{kitti}, M2DGR\cite{m2dgr}, M3DGR\cite{m3dgr}, NTU VIRAL\cite{ntu}, and a self-collected AVIA dataset.
The main information of these datasets are summarized in the following table.

\begin{table}[h!]
   \centering
   \caption{Main Information of the Datasets Used in Experiments.}
   \setlength{\tabcolsep}{7pt}
   \begin{tabular}{@{}l|llll@{}}
   \toprule
   Dataset     & IMU & LiDAR              & Platform & Scenario \\
   \midrule
   KITTI       & No  & Velodyne HDL-64E   & Vehicle & Urban \\
   M2DGR       & Yes & Velodyne VLP-32C   & UGV     & Outdoor\\
   M3DGR       & Yes & Livox AVIA         & UGV     & Corridor\\
   NTU VIRAL   & Yes & Ouster OS1-16      & UAV     & Outdoor \\
   AVIA        & Yes & Livox AVIA         & Hand    & Outdoor \\
   \bottomrule
   \end{tabular}
   \label{tab:datasets}
\end{table}

\begin{table*}[h]
   \centering
   \caption{Accuracy (ATE in meters) Comparison on KITTI Odometry Training Sequences}
   \setlength{\tabcolsep}{8.5pt}
   \begin{tabular}{@{}lcccccccccccc@{}}
   \toprule
   Sequence       & 00       & 01       & 02       & 03       & 04               & 05       & 06       & 07       & 08              & 09       & 10       & Average \\
   (Length [m])   & (3724)   & (2453)   & (5067)   & (561)    & (395)            & (2206)   & (1233)   & (695)    & (3223)          & (1705)   & (920)    & (2016)  \\
   \midrule
   Fast-GICP      & 6.023    & 76.751   & 14.003   & 1.311    & 0.580            & 3.302    & 1.628    & 0.752    & 5.145           & 2.333    & 1.799    & 9.237      \\
   KISS-ICP & 4.284 & 20.182 & 7.668 & 0.847 & 0.464 & 2.068 & 0.931 & \textbf{0.438} & 4.001 & 1.981 & 1.778 & 4.598 \\
   Fast-LIO2      & 3.666    & 19.583   & 9.047    & 0.969    & 0.487            & 1.855    & 0.935    & 0.827    & 3.824           & 2.028    & 1.996    & 4.698      \\
   Faster-LIO     & 6.374    & 19.441   & 9.136    & 0.887    & 0.506            & 1.899    & 0.904    & 0.626    & 3.741           & 1.831    & 1.848    & 5.179      \\
   C3P-VoxelMap   & 3.149    & 4.499    & 8.169    & 0.926    & \textbf{0.231}   & 1.467    & 0.413    & 0.701    & 2.359           & 1.979    & 0.819    & 3.300      \\
   VoxelMap       & 2.079    & 5.140    & 9.789    & 0.850    & 0.258            & 1.007    & 0.401    & 0.503    & \textbf{2.143}  & 1.449    & 0.938    & 3.304      \\
   \midrule
   Ours (w/o \ref{sec:check})  & $\underline{2.022}$ & $\underline{\textbf{3.624}}$ & $\underline{6.025}$ & $\underline{0.827}$ & 0.286 & 1.085 & $\underline{0.390}$ & 0.705 & 2.603 & $\underline{\textbf{1.283}}$ & $\underline{\textbf{0.732}}$ & $\underline{2.574}$ \\
   Ours (full)                 & $\underline{\textbf{1.967}}$    & \underline{3.764}          & $\underline{\textbf{5.912}}$ & $\underline{\textbf{0.790}}$ & 0.291 & $\underline{\textbf{0.997}}$ & $\underline{\textbf{0.385}}$ & 0.460 & 2.795 & \underline{1.373} & \underline{0.735} & $\underline{\textbf{2.570}}$ \\
   \bottomrule
   \label{tab:kitti}
   \end{tabular}
   \vspace{-0.35cm}
   \flushleft \footnotesize{\vspace{0.01cm} Bolded results indicate the best performance among all methods; underlined values indicate our method surpassing all comparative methods.}
   \vspace{-0.5cm}
\end{table*}

The comparative methods employed in our study include Fast-LIO2\cite{fast-lio2}, Faster-LIO\cite{faster-lio}, Fast-GICP\cite{fast-gicp}/IG-LIO\cite{ig-lio}, KISS-ICP\cite{kiss-icp}, VoxelMap\cite{voxelmap}, and C3P-VoxelMap\cite{c3p-voxelmap}.
These state-of-the-art Lidar-Inertial Odometry (LIO) systems are known for their remarkable accuracy and robustness, and have been widely adopted in previous research.
Fast-LIO2\cite{fast-lio2} and Faster-LIO\cite{faster-lio} are point cloud map based LIO systems, utilizing ik-d tree and voxel-based map structures respectively.
GICP \cite{gicp} models local point cloud geometry with Gaussian distributions and uses the Mahalanobis distance for distribution-to-distribution matching.
IG-LIO \cite{ig-lio} integrates GICP into the IEKF framework, where the environment is discretized into voxels, with the points in each voxel modeled as a Gaussian.
KISS-ICP \cite{kiss-icp} achieves accurate and robust registration through an adaptive threshold for data association.
VoxelMap\cite{voxelmap} and its variant\cite{c3p-voxelmap} represent the environment with probabilistic planes and build residuals by matching points to these probabilistic planes.

To ensure a fair comparison, all odometry methods are tested using the same preprocessing pipeline and their default parameters.
For unified parameters, specifically voxel downsampling size and the maximum number of iterations, we enforce identical settings across all methods: 0.5m and 3, respectively, for general datasets, and 0.1m and 5 for the degraded corridor sequences.
VoxelMap and its variants (including the proposed method) are configured with the same voxel-related parameters, using a maximum voxel size of 3m for general datasets and 0.5m for the degraded corridor datasets, and a maximum octree layer level of 4 for all datasets.
All experiments are conducted on a laptop equipped with an Intel i9-13900HX CPU and 32GB RAM.

\subsection{Accuracy on KITTI}
\label{sec:kitti}


In this section, we evaluate our method on the KITTI odometry dataset, which is a large-scale LiDAR-only dataset in urban scenarios, with each frame pre-corrected for motion distortion.
Since the KITTI dataset does not provide IMU data, all IEKF-based LIO methods adopt the same constant-velocity motion model as the prior to ensure fairness.
The localization accuracy is evaluated using the root mean square error (RMSE) of the Absolute Trajectory Error (ATE) after trajectory alignment, where a smaller ATE value corresponds to higher localization accuracy.
The detailed comparison results are shown in Table~\ref{tab:kitti}.
The last column is computed as a frame-count-weighted average over all sequences.

Results show that our method (Ours (full)) achieves the best overall accuracy among all compared algorithms, with an average ATE of 2.570 meters.
Compared to the current state-of-the-art methods, our approach yields a relative improvement of over 20\%.
Among the 11 sequences, Ours (w/o plane validity check) achieves the best performance on 7 sequences, while Ours (full) achieves the best on 8 sequences,
demonstrating superior accuracy compared to all baseline methods.

The improvement primarily arises from the geometry-driven recursive map construction and the distribution-based plane validity check.
The former substantially reduces the influence of outliers and produces more accurate plane representations,
while the recursive reuse of identified outliers preserves valuable geometric information.
Moreover, this strategy significantly reduces the over-segmentation of single physical planes, which is common in original VoxelMap under strict thresholds, thereby reducing parameter uncertainty and accelerating convergence.
The latter further improves odometry accuracy by preventing different physical planes being incorrectly fitted as one, thereby avoiding inaccurate parameter fitting and faulty point-to-plane matching.

\subsection{Accuracy on LIO datasets}
For further validation, we conducted additional comparative experiments on four LIO datasets that differ from the KITTI urban scenarios.
In the M2DGR dataset, we use the street sequences (street\_01 to street\_04), which primarily contain outdoor campus street scenes.
In the M3DGR dataset, we employ the corridor sequences, which include degraded corridor scenarios (corr\_1 and corr\_2).
The NTU VIRAL dataset, collected by a UAV, features significant vertical (Z-axis) motion and represents outdoor environments rich in structural features (spms\_03).
%
The self-collected AVIA dataset was recorded using a small field-of-view (FOV) LiDAR with non-repetitive scanning.
It includes outdoor scenarios with both structured and unstructured environments, and was recorded in closed-loop trajectories with coinciding start and end locations.
Since per-frame ground truth is unavailable for M3DGR and AVIA datasets, we evaluate odometry accuracy using end-to-end drift error.
The comparison results are shown in Table~\ref{tab:lio} and Table~\ref{tab:avia}.

\begin{table}[h]
\centering
\captionsetup{justification=centering}
\caption{Accuracy (ATE in meters) on M2DGR \& NTU VIRAL Dataset}
\setlength{\tabcolsep}{5pt} 
\begin{tabular}{@{}lccccc@{}}
\toprule
Sequence     & street\_01 & street\_02 & street\_03 & street\_04 & spms\_03 \\
(Length [m]) & (752)      & (1485)     & (424)      & (840)      & (312)    \\
\midrule
Fast-LIO2      & 0.281 & 2.815 & 0.184 & 0.452 & 0.233 \\
Faster-LIO     & 0.358 & 3.043 & 0.177 & 0.635 & 0.431 \\
IG-LIO         & 0.364 & 3.184 & 0.188 & 0.485 & 0.223 \\
C3P-VoxelMap   & 0.378 & 3.260 & 0.131 & 1.154 & 1.200 \\
VoxelMap       & 0.352 & 2.516 & 0.133 & 1.273 & 0.215 \\
\midrule
Ours (w/o \ref{sec:check}) & 0.295 & 2.654 & $\underline{\textbf{0.130}}$ & $\underline{0.397}$ & $\underline{\textbf{0.205}}$ \\
Ours (full) & $\underline{\textbf{0.274}}$ & $\underline{\textbf{2.061}}$ & $\underline{\textbf{0.130}}$ & $\underline{\textbf{0.355}}$ & $\underline{0.206}$ \\
\bottomrule
\end{tabular}
\vspace{-0.5cm} 
\label{tab:lio}
\end{table}

The results show that our method achieves excellent localization accuracy across all sequences in M2DGR.
Ours (full) achieves the highest number of best results and consistently outperforms all baseline methods.
This strong performance arises from VoxelMap's inherent adaptability to diverse environments and the precise plane fitting and matching achieved through our recursive map construction with distribution-based plane validity check, as discussed in Sec.~\ref{sec:kitti}.
In structured outdoor scenarios (spms\_03), our method achieves slightly better localization accuracy compared to VoxelMap.
This is because such environments typically contain abundant, well-organized plane structures (e.g., walls) with few outliers.
As a result, the planes fitted by our method are largely consistent with those produced by VoxelMap.
In contrast, in general street environments (street\_01 to street\_04), our method outperforms other approaches more significantly.
This improvement is mainly attributed to the more accurate plane fitting achieved by our method in scenarios with a high proportion of outliers or unstructured elements.

\begin{table}[h!]
\centering
\captionsetup{justification=centering} 
\caption{End-to-End Error (in meters) on M3DGR \& AVIA Datasets}
\setlength{\tabcolsep}{6pt} 
\begin{tabular}{@{}lccccccc@{}}
\toprule
Method       & corr1     & corr2     & avia1 & avia2 & avia3 & avia4 \\
(Length [m]) & (330)     & (236)     & (923) & (514) & (572) & (500) \\
\midrule
Fast-LIO2    & 10.540     & 11.586     & 3.195 & 0.990 & 2.649 & 0.123 \\
Faster-LIO   & --         & 14.140     & 1.514 & 2.512 & 1.039 & 0.079 \\
C3P-VoxelMap & 15.130    & \textbf{1.843}     & 3.964 & 3.796 & 0.181 & 0.185 \\
VoxelMap     & 6.104     & 14.630     & 2.412 & 0.039 & 0.133 & 0.027 \\
\midrule
Ours(full)   & \textbf{1.036} & 2.518 & \textbf{0.014} & \textbf{0.013} & \textbf{0.124} & \textbf{0.019} \\
\bottomrule
\end{tabular}
\vspace{-0.1cm}
\label{tab:avia}
\end{table}

\begin{figure}[t]
   \vspace*{0.5\baselineskip} 
   \centering
   \includegraphics[width=0.48\textwidth]{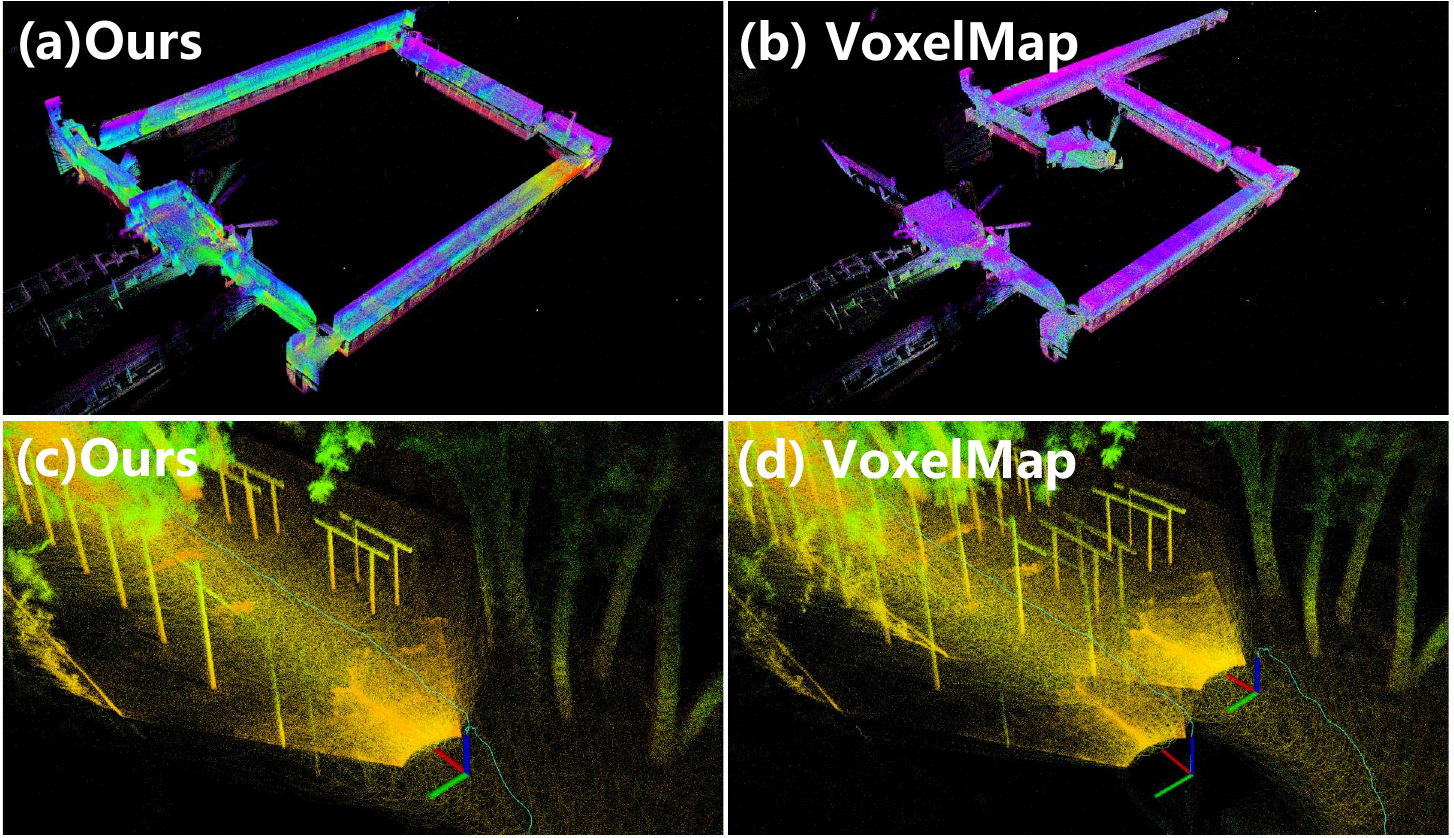}
   \caption{
      (a) and (b) Comparison of mapping results between the proposed method and VoxelMap on the M3DGR dataset (corridor2 sequence).
      (c) and (d) End-to-end error comparison between the proposed method and VoxelMap on the AVIA dataset (avia1 sequence).
      The proposed method achieves better mapping quality and localization accuracy.
   }
   \label{fig:avia}
   \vspace{-0.5cm}  
\end{figure}

This is further validated on the M3DGR and self-collected AVIA datasets, representing corridor-degraded environments and outdoor scenes with abundant unstructured features (e.g., trees).
The results show that our method maintains stable advantages across multiple sequences, particularly in the corr1 corridor scenario and avia1 outdoor sequence, demonstrating that the proposed method can still achieve high localization accuracy even in challenging and degenerate environments.
As the visualization results shown in Fig.~\ref{fig:avia}, VoxelMap exhibits point cloud misalignment caused by accumulated drift.
In contrast, the proposed method produces a more accurate and consistent mapping result.

In summary, compared with current state-of-the-art methods, our method performs well across various scenarios and achieves notably greater improvements in localization accuracy, particularly in long-range street and urban environments as well as in several degraded scenarios.

\subsection{Runtime and Memory Usage}\label{sec:runtime}
To evaluate the computational efficiency and memory usage of our method,
we compare our method against VoxelMap on the KITTI dataset in terms of runtime and memory consumption.
For fairness, the same LRU cache mechanism is applied to VoxelMap, and both methods use identical multi-threading settings during the IEKF update process.
Runtime is computed as the average processing time per frame for each individual sequence, while memory usage refers to the total memory consumption at the end of each sequence.
The detailed results are presented in Table~\ref{tab:runtime}.
\begin{table}[h!]
   \centering
   \caption{Runtime and Memory Usage on KITTI Datasets}
   \begin{tabular}{@{}lcccc@{}}
   \toprule
   \multirow{2}{*}[-0.5ex]{\begin{tabular}{c}Sequence\\(Length(m))\end{tabular}} & \multicolumn{2}{c}{VoxelMap} & \multicolumn{2}{c}{Ours (full)} \\
   \cmidrule(lr){2-3} \cmidrule(lr){4-5}
   & Time (ms) & Mem (GB) & Time (ms) & Mem (GB) \\
   \midrule
   00 (3724)  & 44.37 & 4.09 & \textbf{40.18} & \textbf{3.16} \\
   01 (2453)   & \textbf{64.52} & 2.55 & 75.23 & \textbf{2.37} \\
   02 (5067)  & 45.94 & 4.53 & \textbf{44.44} & \textbf{4.08} \\
   03 (561)  & 62.86 & 1.06 & \textbf{58.52} & \textbf{0.93} \\
   04 (395)  & \textbf{60.10} & 0.60 & 64.98 & \textbf{0.57} \\
   05 (2206) & 48.17 & 2.77 & \textbf{44.70} & \textbf{2.03} \\
   06 (1233)  & 73.64 & 1.44 & \textbf{62.77} & \textbf{1.04} \\
   07 (695)  & 44.01 & 1.05 & \textbf{40.90} & \textbf{0.84} \\
   08 (3223)  & 55.22 & 4.91 & \textbf{53.22} & \textbf{3.86} \\
   09 (1705)  & 54.81 & 2.11 & \textbf{53.86} & \textbf{1.88} \\
   10 (920)  & 42.58 & 1.18 & \textbf{40.93} & \textbf{0.99} \\
   \bottomrule
   \label{tab:runtime}
   \end{tabular}
   \vspace{-0.5cm}
\end{table}

\begin{table*}[t]
   \centering
   \caption{Module-wise Runtime on KITTI Sequence 00 (ms)}
   \setlength{\tabcolsep}{8pt} 
   \renewcommand{\arraystretch}{1.1} 
   \begin{tabular}{c|ccccc|c|c|c|c}
   \hline
   \multirow{2}{*}{} & \multicolumn{6}{c|}{Map Maintenance} & \multirow{2}{*}[-0.5ex]{IEKF Update} & \multirow{2}{*}[-0.5ex]{Other} & \multirow{2}{*}[-0.5ex]{Total} \\
   \cline{2-7}
    & plane param update & convince update & RANSAC & plane check & other & total & & & \\
   \hline
   Voxelmap & 2.34 & 10.71 & - & - & 2.72 & 15.77 & 16.46 & 12.02 & 44.25 \\
   Ours & 1.95 & 8.65 & 4.19 & 0.29 & 2.15 & 17.23 & 11.72 & 11.23 & 40.18 \\
   \hline
   \end{tabular}
   \label{tab:runtime_each}
   \vspace{-0.5cm}
\end{table*}
The results show that our method consistently achieves lower computational time and memory usage than VoxelMap across nearly all sequences.
Through a detailed analysis of the runtime of each module (see Table~\ref{tab:runtime_each}),
although RANSAC introduces a certain amount of additional computation, we compensate for this overhead through optimizations in other modules.
Our method can extract large planar structures at earlier stages, allowing their parameter uncertainties to converge more quickly. This reduces the frequency of expensive covariance-update operations.
In addition, during IEKF update, we eliminate unnecessary variable passing, which further reduces both runtime and memory usage. Consequently, the total computational cost of our method remains comparable to that of VoxelMap.

\subsection{Stability Analysis}
Due to the randomness introduced by RANSAC, we conducted a stability experiment to assess the consistency of our algorithm's localization accuracy and computational efficiency under different random seeds.
Specifically, we performed independent experiments using five different random seeds on the KITTI\_00 and KITTI\_01 sequences, recorded ATE and runtime for each trial.

\begin{table}[h!]
\centering
\captionsetup{justification=centering} 
\caption{Accuracy and Runtime under Different Random Seeds on KITTI}
\setlength{\tabcolsep}{5pt}
\begin{tabular}{@{}ccccc@{}}
\toprule
\multirow{2}{*}[-0.6ex]{Seed} & \multicolumn{2}{c}{KITTI 00} & \multicolumn{2}{c}{KITTI 01} \\
\cmidrule(lr){2-3} \cmidrule(lr){4-5}
 & ATE (m) & Time (ms) & ATE (m) & Time (ms) \\
\midrule
42 & 1.967 & 40.18 & 3.765 & 75.78 \\
1  & 1.982 & 40.24 & 3.825 & 75.49 \\
2  & 1.923 & 40.20 & 3.883 & 76.09 \\
3  & 2.050 & 40.19 & 3.679 & 75.67 \\
4  & 1.848 & 40.23 & 3.804 & 76.14 \\
\midrule
Mean (std) & \textbf{1.954} (0.075) & \textbf{40.21} (0.03) & \textbf{3.791} (0.076) & 75.83 (0.28) \\
\midrule
VoxelMap & 2.079 & 44.37 & 5.140 & \textbf{64.52} \\
\bottomrule
\end{tabular}
\vspace{-0.3cm}
\label{tab:kitti_seed}
\end{table}

As shown in Table~\ref{tab:kitti_seed}, the proposed method achieves consistently better localization accuracy than the original VoxelMap across different random seeds.
The runtime remains largely consistent across different seeds, which is in agreement with the results presented in Section~\ref{sec:runtime}.

\section{Conclusions}
This paper proposes R-VoxelMap, a novel accurate voxel mapping method for LiDAR-based odometry.
Compared to VoxelMap and its variants, R-VoxelMap does not directly fit planes using all points within a voxel during the plane extraction stage.
Instead, it employs a geometry-driven recursive map construction framework with an outlier detect-and-reuse pipeline and a distribution-based plane validity check.
The proposed method effectively mitigates the impact of outliers on plane fitting and prevents the erroneous merging of different physical planes.
Experimental results on diverse datasets,
covering environments with both structured and unstructured sensors,
demonstrate that compared to current state-of-the-art methods, R-VoxelMap effectively enhances localization accuracy without introducing additional processing time or memory overhead.

Although our method effectively mitigates several issues such as the influence of outliers, the VoxelMap-based framework still suffers from high sensitivity to multiple parameters.
In the future, we plan to further investigate strategies to reduce the system's parameter sensitivity and to enhance localization accuracy across diverse environments.

\bibliography{ref}
\bibliographystyle{IEEEtran}

\end{document}